\documentclass[journal]{IEEEtran}
\usepackage{amsmath,amsfonts}
\usepackage{algorithmic}
\usepackage{algorithm}
\usepackage{array}
\usepackage[caption=false,font=normalsize,labelfont=sf,textfont=sf]{subfig}
\usepackage{textcomp}
\usepackage{stfloats}
\usepackage{url}
\usepackage{verbatim}
\usepackage{graphicx}
\usepackage{balance}
\usepackage{cite}
\usepackage{multirow}
\usepackage{multicol}
\usepackage{newfloat}
\usepackage{listings}
\usepackage{utfsym}
\usepackage{MnSymbol}
\usepackage{makecell}
\usepackage{pifont}
\usepackage{booktabs}
\usepackage{enumitem}
\usepackage[colorlinks,
            linkcolor=blue,       
            anchorcolor=blue,  
            citecolor=blue,       
            ]{hyperref}
\usepackage{orcidlink}
\usepackage{soul}

\hypersetup{
    colorlinks=true,
    linkcolor=black,
    filecolor=black,      
    urlcolor=black,
    citecolor=blue,
}
\usepackage[table]{xcolor}
\usepackage{colortbl}

\usepackage{fontawesome5} 
\usepackage{xcolor}       

\usepackage{threeparttable} 

\definecolor{skyblue}{RGB}{198, 226, 255}

\definecolor{myblue}{RGB}{235, 245, 255} 
\definecolor{cmarkgreen}{RGB}{0, 150, 0} 
\definecolor{xmarkred}{RGB}{200, 0, 0}   

\begin{document}

\title{An Efficient and Effective Watermarking Scheme for the Protection of the Intellectual Property Rights of Video Generative Models}

\author{
    Wenhong Huang\raisebox{0.5ex}{\orcidlink{0009-0004-3612-7959}}, 
    Jianwei Fei\raisebox{0.5ex}{\orcidlink{0000-0002-1243-3909}}, 
    Benedetta Tondi\raisebox{0.5ex}{\orcidlink{0000-0002-7518-046X}}, 
    Bin Ma\raisebox{0.5ex}{\orcidlink{0000-0002-9030-7393}}, 
    and Fangjun Huang\raisebox{0.5ex}{\orcidlink{0000-0002-3098-3373}}

    \thanks{
        This work was supported in part by the National Natural Science Foundation of China under Grants 62472454, U2336208, and 62272255, by the Shenzhen Science and Technology Program under Grant JCYJ20250604175534044, and by the Key Laboratory of Computing Power Network and Information Security, Ministry of Education under Grant 2024ZD022. \textit{(Corresponding author: Fangjun Huang.)}
    }

    \thanks{
        Wenhong Huang and Fangjun Huang are with the School of Cyber Science and Technology, Shenzhen Campus of Sun Yat-sen University, Shenzhen 518107, China, and also with Guangdong Provincial Key Laboratory of Information Security Technology, Guangzhou 510006, China 
        (e-mail: \nolinkurl{huangwh79@mail2.sysu.edu.cn};
        \nolinkurl{huangfj@mail.sysu.edu.cn}).
        Jianwei Fei is with the Department of Information Engineering, University of Florence, Florence 50139, Italy.
        Benedetta Tondi is with the Department of Information Engineering and Mathematics, University of Siena, Siena 53100, Italy.
        Bin Ma is with the Key Laboratory of Computing Power Network and Information Security, Ministry of Education, Qilu University of Technology (Shandong Academy of Sciences), Jinan 250353, China.
        } 

}


\maketitle

\begin{abstract}
The rapid development of video generative models (VGMs) has enabled the generation of highly realistic synthetic videos, raising concerns about the intellectual property rights (IPR) of these models. In particular, two closely related forensic tasks remain largely unaddressed: \textit{synthetic video verification} (determining whether a video was generated by a protected VGM) and \textit{model ownership verification} (determining whether a suspect VGM is an unauthorized copy of a protected VGM). In this paper, we propose a new in-generation watermarking scheme that can address the two verification tasks. First, a novel video watermarking network named VidMark is presented, which incorporates a two-scale discrete wavelet transform (DWT) decomposition and a global temporal attention block (GTAB) to enhance watermark robustness and imperceptibility. Second, we present a decoder-guided fine-tuning procedure. By leveraging the frozen VidMark decoder, this process enables VGMs to synthesize videos carrying an imperceptible, robust, and model-specific watermark. Finally, two verification frameworks are established to perform synthetic video verification and model ownership verification. Extensive experiments on representative VGMs demonstrate that the proposed scheme achieves over 99\% watermark extraction accuracy and 100\% verification accuracy on both tasks, with negligible impact on video generation quality. Furthermore, the watermarks exhibit strong robustness against a comprehensive range of video-level and model-level attacks.
\end{abstract}

\begin{IEEEkeywords}
  Watermarking,
  video generative models,
  ownership verification,
  information forensics,
  deep learning security.
\end{IEEEkeywords}

\maketitle

\section{Introduction}
\label{sec1_introduction}

\IEEEPARstart{V}{ideo} generative models (VGMs) are a class of deep neural network (DNN) models that can synthesize videos from text, images, or other inputs. In recent years, large-scale VGMs \cite{blattmann2023stable,karras2023dreampose,polyak2024movie,zheng2024open,wan2.1} based on the video latent diffusion model~\cite{blattmann2023align} (video LDM) architecture have attracted widespread attention from both academia and industry. This is largely due to their ability to generate realistic, high-resolution videos while performing video modeling in a computationally efficient compressed latent space. Despite these advances, the high training cost and commercial value of VGMs raise pressing intellectual property concerns. On the one hand, generated videos may be misappropriated and redistributed without authorization. On the other hand, the VGM itself may be stolen, copied, and redeployed. These risks motivate two closely related forensic tasks: (i) \textit{synthetic video verification}, i.e., determining whether a single video of unknown source was generated by a protected VGM, and (ii) \textit{model ownership verification}, i.e., determining whether a suspect VGM of unknown origin is a stolen copy of a protected VGM, so as to safeguard the intellectual property rights (IPR) of model owners.

Currently, watermarking is widely adopted for protecting the IPR of DNN models, especially generative models~\cite{fei2022supervised,fernandez2023stable,kim2024wouaf,feng2024aqualora,wen2024tree,fei2024wide,liu2025implanting,min2024watermark,tang2024robust,hu2025deep}. It enables a deployed generative model and/or its outputs to be linked to the legitimate owner, supporting both ownership verification and the tracing of unauthorized use~\cite{barni2021dnn}. 
Notably, in-generation watermarking stands as the sole paradigm capable of simultaneously addressing synthetic video verification and model ownership verification. By integrating the watermark embedding into the content generation process, it ensures that every generated output carries a model-specific watermark~\cite{fernandez2023stable}, thereby establishing a binding link between the watermark, the output, and the model. For image generative models (IGMs), a common approach to in-generation watermarking involves fine-tuning model parameters guided by pre-trained watermark decoders, such as HiDDeN~\cite{zhu2018hidden}. These methods have demonstrated good performance in terms of image fidelity, watermark robustness, and embedding capacity across different IGMs. However, this paradigm is difficult to apply directly to VGMs for several reasons. First, the higher dimensionality and more complex temporal modeling of video data lead to substantially higher training and fine-tuning costs~\cite{blattmann2023align}. Second, watermarks in the video domain must withstand temporal attacks such as video compression and frame averaging, which are unique to video and have no direct counterparts in image watermarking. Third, as VGMs typically generate high-resolution videos (e.g., 512$\times$512 or higher) with significant computational overhead, the ideal video watermark decoder must be both lightweight and capable of high-resolution processing. However, this requirement is not met by current methods: REVMark~\cite{zhang2023novel} and DVMark~\cite{luo2023dvmark} are limited to low-resolution videos (e.g., 128$\times$128), while VideoSeal~\cite{fernandez2024video}, though effective, demands substantial computational resources (e.g., 16 GPUs for training).

To this end, we propose a new in-generation watermarking scheme for VGMs. The core of this scheme is VidMark, a lightweight encoder-decoder network specifically designed for high-resolution videos. It employs a two-scale discrete wavelet transform (DWT) ~\cite{mallat1989theory} decomposition to embed watermarks into the low-frequency components of video frames, which can improve the watermark robustness against common geometric/non-geometric distortions~\cite{bhowmik20122d} and expand the network's effective receptive field for high-resolution inputs. Furthermore, a global temporal attention block (GTAB) is specifically designed to capture long-range temporal dependencies between frames. By enforcing temporal consistency in the feature domain, GTAB effectively reduces inter-frame artifacts, thereby improving the overall visual quality of the watermarked video. Once VidMark is fully trained for video watermark embedding and extraction, we freeze its decoder and use it to fine-tune the last few layers of a selected VGM. Guided by a composite loss function, these output-proximate layers of the VGM will learn to replicate the watermark embedding behavior of the VidMark encoder. This process can effectively embed an imperceptible watermark into the generated video while maintaining the original visual fidelity. During the verification phase, extracting the watermark from a given video allows us to ascertain its synthetic origin. Furthermore, by gathering multiple videos generated from a suspect model and extracting watermarks from them, we can determine whether the suspect model originates from a protected VGM, enabling model ownership verification.

Our evaluations show that VidMark itself achieves 100\% extraction accuracy for a 128-bit watermark embedded in 512$\times$512 videos, while maintaining high visual fidelity. Furthermore, it is highly efficient to train, requiring only 5.89 GB of peak GPU memory. This is a significant advantage over existing methods: VideoSeal reported training on 16 GPUs~\cite{fernandez2024video}, and REVMark~\cite{zhang2023novel} requires 17 GB of peak GPU memory. Applying the pre-trained VidMark decoder to watermark several widely used text-to-video and image-to-video VGMs (including Stable Video Diffusion (SVD)~\cite{blattmann2023stable}, Open-Sora~\cite{zheng2024open}, and Wan~\cite{wan2.1}) demonstrates that the new scheme significantly outperforms baselines such as Stable Sig.~\cite{fernandez2023stable} and WOUAF~\cite{kim2024wouaf} in terms of visual fidelity, verification accuracy, and watermark robustness.

The remainder of this paper is organized as follows. Section~\ref{sec2_related_work} reviews recent watermarking schemes. Section~\ref{sec2_proposed_watermarking_scheme} presents VidMark and the watermarking procedure for VGMs. Section~\ref{sec3_watermark_verification_framework} introduces the verification frameworks for synthetic video and model ownership verification. Sections~\ref{sec4_experiments_on_video_watermarking} and~\ref{sec5_experiments_on_VGM_watermarking} report the experimental results on video watermarking and VGM watermarking, respectively. Section~\ref{sec6_conclusion} concludes the~paper.

\section{Related Work}
\label{sec2_related_work}

Numerous watermarking techniques have been proposed to safeguard generative models, including generative adversarial networks (GANs)~\cite{goodfellow2014generative}, diffusion models (DMs) ~\cite{ho2020denoising}, and latent diffusion models (LDMs) ~\cite{rombach2022high,blattmann2023align}. Given that most of the current large-scale IGMs and VGMs are built upon LDM architectures, this section is dedicated to introducing watermarking schemes for LDM-based IGMs and VGMs. 

Existing watermarking schemes fall into three paradigms based on the stage in which the watermark is embedded: (i)~pre-generation schemes, which embed watermarks into initialization variables (e.g., initial latent) before the diffusion process begins and extract them via inversion; (ii)~post-generation schemes, which embed watermarks into the output after synthesis; (iii)~in-generation schemes, which fine-tune model parameters to embed the watermark during generation.

Among pre-generation schemes for IGMs, Tree-Ring~\cite{wen2024tree} embeds the watermark signal into the frequency domain of the initial latent, while WIND~\cite{arabi2025hidden} augments the initial latent with generated Fourier patterns to embed watermark information. Other methods focus on the initial latent sampling process itself. For example, Gaussian Shading~\cite{yang2024gaussian} encodes watermarks during the sampling of the initial latent, and PRC~\cite{gunn2025prc} embeds watermarks by selecting the initial latent from a pseudorandom codebook. Meanwhile, post-generation schemes embed watermarks into the already synthesized content using separate encoder-decoder networks, following classic multimedia watermarking strategies such as HiDDeN~\cite{zhu2018hidden} and StegaStamp~\cite{tancik2020stegastamp}. Overall, pre-generation and post-generation schemes watermark the content without performing any modification to the model. Consequently, they do not inherently support model ownership verification, as the model remains unprotected~\cite{fernandez2023stable,kim2024risk,yang2024gaussian}. In contrast, in-generation schemes watermark the model directly in such a way that the watermark can be read from the outputs it produces~\cite{fei2022supervised,fernandez2023stable}, providing a unified solution for synthetic image verification and model ownership verification.

Various in-generation watermarking schemes for IGMs have been proposed in recent years. One type of solution is to fine-tune the latent decoder. 
Fei et al.~\cite{fei2022supervised} first introduce supervised fine-tuning for watermarking generative models such that the outputs intrinsically carry an invisible signature.
A recent work, Stable Sig. ~\cite{fernandez2023stable}, fine-tunes the LDM's latent decoder (i.e., VAE decoder) to ensure that the generated image carries a signature decodable by a fixed extractor (HiDDeN~\cite{zhu2018hidden}). WOUAF~\cite{kim2024wouaf} employs weight modulation to fine-tune the latent decoder, thereby embedding traceable fingerprints for distinct users. The second type of solution incorporates the watermark into the diffusion process. WaDiff~\cite{min2024watermark} introduces watermark bits as conditional inputs to the diffusion network, supervising the image generation with a watermark extraction objective. AquaLoRA~\cite{feng2024aqualora} injects a watermark LoRA module into the U-Net of Stable Diffusion~\cite{rombach2022high}.

\begin{table}[t]
    \centering
    \begin{threeparttable}
        \caption{Comparison of representative schemes for watermarking latent diffusion models. 
        }
        \label{tab_method}
        \footnotesize
        \setlength{\tabcolsep}{1.2mm} 
        \renewcommand\arraystretch{1.0}
        
        \begin{tabular}{lcccc}
        \toprule
        \multirow{2}{*}{\textbf{Method}} & \multirow{1}{*}{\textbf{Model}} & \multirow{2}{*}{\textbf{SVV}} & \multirow{2}{*}{\textbf{MOV}} & \textbf{Temp.} \\
        
        
         & \textbf{Arch.} & & & \textbf{Robust.} \\
        \midrule
        
        \multicolumn{5}{l}{\textit{In-generation Watermarking Schemes for IGMs}} \\
        Stable Sig.~\cite{fernandez2023stable} (in-gen.)
            & Image LDMs & \textcolor{black}{\ding{51}}$^\dagger$ & \ding{51}$^\dagger$ & \textcolor{xmarkred}{\ding{55}} \\ 

        WOUAF~\cite{kim2024wouaf} (in-gen.)
             & Image LDMs & \textcolor{black}{\ding{51}}$^\dagger$ & \ding{51}$^\dagger$ & \textcolor{xmarkred}{\ding{55}} \\

        \midrule
        \multicolumn{5}{l}{\textit{Watermarking Schemes for VGMs}} \\
        
        VideoShield~\cite{hu2025videoshield} (pre-gen.) & Video LDMs & \ding{51} & \textcolor{xmarkred}{\ding{55}} & \ding{51} \\

        VideoMark~\cite{hu2025videomark} (pre-gen.)
            & Video LDMs & \ding{51} & \textcolor{xmarkred}{\ding{55}} & \ding{51} \\

        SVS~\cite{liu2025implanting} (in-gen.)
            & \textcolor{xmarkred}{SVD}~\cite{blattmann2023stable} & \ding{51} & \ding{51} & \ding{51} \\
            
        \midrule
        \rowcolor{gray!8}
        \textit{\textbf{Our Scheme}} (in-gen.)
            & \textcolor{black}{General VGMs}
            & \textbf{\textcolor{black}{\ding{51}}} &  \textbf{\textcolor{black}{\ding{51}}} &  \textbf{\textcolor{black}{\ding{51}}} \\

        \rowcolor{gray!8}
            \quad \ensuremath{\hookrightarrow} \textit{\textbf{VidMark}}$^\ddagger$ (post-gen.) & Videos & \textbf{\textcolor{black}{\ding{51}}} & \textbf{\textcolor{black}{\ding{55}}} &  \textbf{\textcolor{black}{\ding{51}}}\\
            
        \bottomrule
        \end{tabular}
        
        \vspace{5pt}
        
        \begin{tablenotes}
            \footnotesize
            \item[1] \textbf{Model Arch.} (Model Architecture) indicates the model architecture of targeted IGMs/VGMs.
            \item[2] \textbf{SVV} refers to the \textit{synthetic video verification} defined in Section~\ref{sec1_introduction}. 
            \item[3] \textbf{MOV} represents the \textit{model ownership verification} defined in Section~\ref{sec1_introduction}. 
            \item[4] \textbf{Temp. Robust.} (Temporal Robustness) reflects the capability to resist video-specific attacks (e.g., motion blurring).
            \item[5] {in-gen.} (in-generation); {pre-gen.} (pre-generation); {post-gen.} (post-generation)
            \item[$\dagger$] indicates that the method can be applied to the verification tasks frame by frame.
            \item[$\ddagger$] \textbf{\textit{VidMark}} serves as the core component within our full scheme but can also be deployed as a standalone post-generation solution.
        \end{tablenotes}
    \end{threeparttable}
\end{table}

In contrast, research on watermarking for VGMs remains limited. Most existing solutions follow a pre-generation paradigm. For instance, VideoShield~\cite{hu2025videoshield} derives template bits from the watermark and samples the initial noise accordingly to produce a watermarked initial latent, while VideoMark~\cite{hu2025videomark} initializes the latent with a pseudorandom message sequence to encode the watermark. Similar to pre-generation IGM watermarking, these methods are typically designed for synthetic video verification and lack native support for model ownership verification. As for in-generation schemes for VGMs, to the best of our knowledge, the only published work to date is SVS~\cite{liu2025implanting}, which embeds watermarks by fine-tuning the deflickering decoder of the SVD model~\cite{blattmann2023stable} via model distillation. However, this approach is heavily dependent on the specific architecture (e.g., requiring the SVD deflickering decoder structure);
consequently, its applicability remains narrow in the face of rapidly diversifying VGMs.

Table~\ref{tab_method} presents a comparison of representative watermarking schemes. As shown, for general VGMs, there is no solution that supports both synthetic video verification and model ownership verification while maintaining temporal robustness. The existing in-generation schemes for IGMs only support frame-by-frame synthetic video verification and model ownership verification, lacking temporal robustness~\cite{luo2023dvmark}; the pre-generation schemes for VGMs fail to safeguard model ownership; and the in-generation scheme, SVS~\cite{liu2025implanting}, is specifically tailored for the SVD model~\cite{blattmann2023stable}, restricting its generalization across other VGMs. In contrast, our approach provides a unified solution, in which the core component, VidMark, also supports flexible deployment as an independent post-generation solution.
\section{Proposed Watermarking Scheme}
\label{sec2_proposed_watermarking_scheme}
This section presents our watermarking scheme for protecting VGMs. Specifically, Section~\ref{sec2_1_VidMark} introduces the architecture and training of the proposed VidMark, as well as the two-scale DWT decomposition and our newly designed GTAB in detail. Then, Section~\ref{sec2_2_watermarking_VGMs} describes how VidMark is used to watermark VGMs, including the video generation process in VGMs and the watermarking pipeline.

\begin{figure*}[tbp]
    \centering
    \includegraphics[width=.95\linewidth]{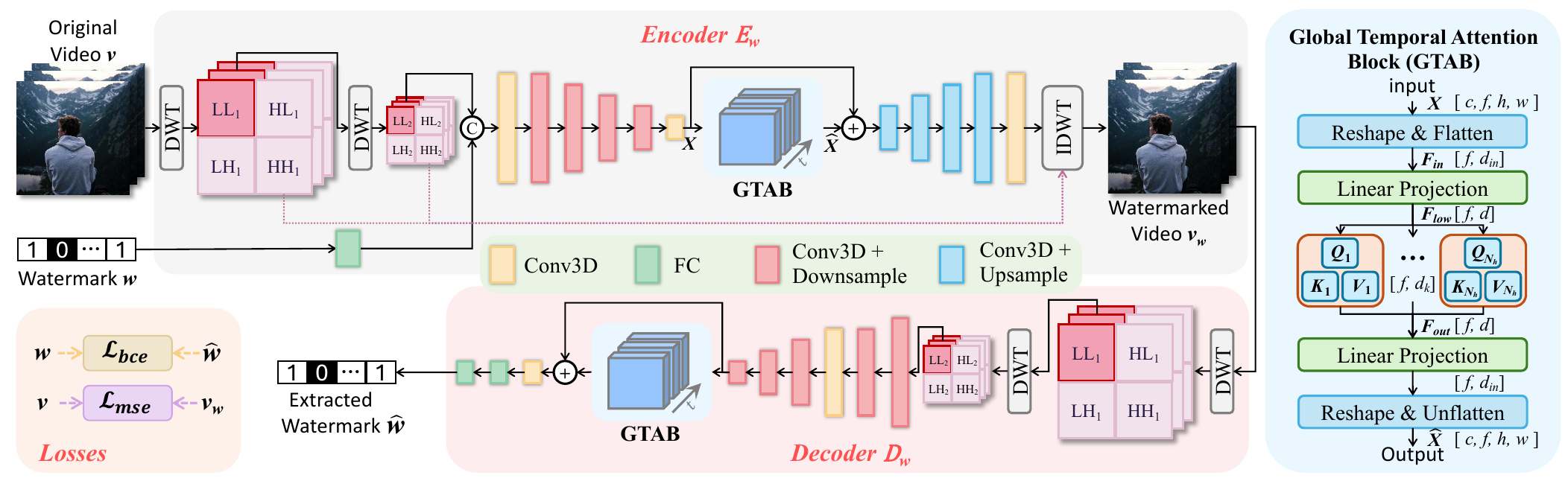}
    \caption{ Framework of the proposed VidMark for video watermarking, including the encoder $E_w$ and decoder $D_w$. The ``Conv3D + Downsample'' block denotes a cascade of two 3D convolutional layers: the first layer simultaneously performs spatial downsampling and channel expansion, while the second refines the features. Conversely, the ``Conv3D + Upsample'' block begins with an upsampling operation to restore spatial resolution, followed by two convolutional layers that fuse features and reduce channel dimensions. For visual clarity, the skip connections between downsampling and upsampling blocks, as well as the normalization layers (e.g., group normalization, layer normalization), are omitted.
    }
\label{Fig_vidmark}
\end{figure*}

\subsection{VidMark}
\label{sec2_1_VidMark}
Compared to image watermarking, video watermarking needs to handle temporal synchronization and motion-related issues such as frame averaging and motion blurring, which makes it more challenging to simultaneously achieve robustness and imperceptibility of the watermarks~\cite{Lin2004temporal}. To this end, we propose VidMark, which is designed to effectively model the spatiotemporal dependencies inherent in videos, thereby enabling both robust and imperceptible watermark embedding. As shown in Fig.~\ref{Fig_vidmark}, VidMark consists of an encoder $E_w$ for watermark embedding and a decoder $D_w$ for watermark extraction. The VidMark encoder $E_w$ employs U-Net~\cite{ronneberger2015u} as its backbone network, with 3D convolutional (Conv3D) blocks as its basic building units. Its feature representation capability is further augmented by incorporating a two-scale DWT decomposition and the newly designed GTAB. The VidMark decoder $D_w$ includes the two-scale DWT decomposition, the GTAB, and a series of downsampling Conv3D and fully connected layers for extracting the embedded watermark from a watermarked video.

\subsubsection{VidMark Encoder}
The inputs of the VidMark encoder $E_w$ include a randomly generated $k$-bit watermark vector $w \in \{0,1\}^k$ and an input video $v \in \mathbb{R}^{C \times F \times H \times W}$, where $C$, $F$, $H$, and $W$ represent the dimensionality of the video in terms of channels, frames, height, and width, respectively. The 2D DWT~\cite{mallat1989theory} is performed on each frame of the input video $v$, decomposing it into four subbands: (a) the approximation/low-low subband (LL$_{1}$), (b) the horizontal detail/low-high subband (LH$_{1}$), (c) the vertical detail/high-low subband (HL$_{1}$), and (d) the diagonal detail/high-high subband (HH$_{1}$). Given that embedding watermarks in the low-low subband generally provides higher robustness~\cite{jang2024waterf,huang2025new}, we employ a two-scale DWT strategy. In this approach, LL$_{1}$ is further decomposed into four coarser subbands: LL$_{2}$, LH$_{2}$, HL$_{2}$, and HH$_{2}$. Among them, only the coarsest low-low subband (LL$_{2}$) is used for watermark embedding, while the other subbands (LH$_{1}$, HL$_{1}$, HH$_{1}$, LH$_{2}$, HL$_{2}$, and HH$_{2}$) are reserved for subsequent video reconstruction via the inverse DWT (IDWT). This multi-scale design can improve the watermark robustness against common signal processing and compression attacks~\cite{bhowmik20122d}. Moreover, the spatial downscaling brought by the two DWT decompositions allows convolution operations at lower scales to capture broader contextual information of the original frame. This facilitates more effective modeling of local structures and makes it suitable for watermarking high-resolution videos.

Meanwhile, the watermark vector $w$ is projected into a message tensor via a fully connected (FC) layer and reshaped to match the spatial and temporal dimensions of LL$_{2}$. Then, this expanded message tensor is concatenated with LL$_{2}$ along the channel dimension. This operation constructs a composite input, allowing the subsequent convolutional layers to fuse the visual content with the watermark signal within a joint feature space. The fused features are then processed through multiple Conv3D and downsampling Conv3D layers to generate a high-dimensional spatiotemporal feature map $X$. To overcome the limited receptive field of convolution and to explicitly model long-range temporal dependencies, $X$ is fed into the GTAB. As detailed in Section~\ref{sec2_1_3_STAB}, the GTAB generates the temporally enhanced feature map $\hat{X}$, which is then fused with $X$ via a residual connection. This design effectively injects global temporal features into local features while preserving the original spatial details, thereby significantly enhancing the model's feature representation capabilities. Subsequently, a series of upsampling Conv3D layers is applied, while skip connections are introduced to link the corresponding downsampling and upsampling layers to preserve detailed features. The last Conv3D layer projects the fused feature map back to the original channel dimension. Finally, the processed feature map is combined with the reserved LH$_{1}$, HL$_{1}$, HH$_{1}$, LH$_{2}$, HL$_{2}$, and HH$_{2}$ subbands and reconstructed into the spatial domain through IDWT to produce the watermarked video $v_w$.

\subsubsection{VidMark Decoder}
The watermarked video $v_w$ is fed into the VidMark decoder $D_w$ to extract the watermark. As shown in Fig.~\ref{Fig_vidmark}, we first apply the two-scale DWT decomposition to each frame of $v_w$ to obtain the corresponding LL$_{2}$ subband. Then, LL$_{2}$ is fed into a series of Conv3D and downsampling Conv3D layers to extract high-dimensional features, which are further processed through the GTAB and successive fully connected layers to obtain the extracted watermark $\hat{w}$.

\subsubsection{GTAB}
\label{sec2_1_3_STAB}
The GTAB, shown on the right of Fig.~\ref{Fig_vidmark}, is proposed to explicitly capture the long-range temporal dependencies between frames. Let $X \in \mathbb{R}^{c\times f\times h\times w}$ denote the input feature map, where $c$, $f$, $h$, and $w$ represent the channel, frame, height, and width dimensions, respectively. First, to perform attention along the temporal dimension, the input feature map $X$ is reshaped and flattened along its spatial and channel dimensions into a sequence of frame features $F_{in}\in \mathbb{R}^{f\times d_{in}}$, where $d_{in}=c\cdot h\cdot w$. Since $d_{in}$ is typically large, we employ a linear layer to project $F_{in}$ into a lower-dimensional feature sequence $F_{low}\in \mathbb{R}^{f\times d}$. Here, $d$ is the Transformer model's internal dimension, defined as $d=N_h\cdot d_k$, where $N_h$ is the number of heads and $d_k$ is the head dimension. Subsequently, multi-head attention~\cite{vaswani2017attention,jain2024peekaboo} with $N_h$ heads is applied to the feature sequence $F_{low}$ to model the global interactions between frames. Specifically, the feature sequence $F_{low}$ is split into $N_h$ independent heads, where the $i$-th head computes its own scaled dot-product attention:
\begin{equation}
    {\mathrm{Head}}_i=\mathrm{softmax}\left(\frac{Q_iK_i^\top}{\sqrt{d_k}}\right)V_i, i\in\{1,\ldots,N_h\}
\end{equation}
where $Q_i,K_i,V_i\in \mathbb{R}^{f\times d_k}$ are the query, key, and value matrices for the $i$-th head. The outputs from all heads are then concatenated and fused via a linear transformation $W_O \in \mathbb{R}^{d\times d}$ to generate the aggregated temporal features: 
\begin{equation}
    F_{out}=\mathrm{Concat}\left({\mathrm{Head}}_1,\ldots,{\mathrm{Head}}_{N_h}\right)W_O\in R^{f\times d}.
\end{equation}
Finally, $F_{out}$ is projected back to the original high-dimensional space $\mathbb{R}^{f\times d_{in}}$ via a linear layer and reshaped to $\hat{X}\in \mathbb{R}^{c\times f\times h\times w}$.
\subsubsection{Losses}
VidMark is designed to generate watermarked videos with high visual fidelity while enabling accurate watermark extraction. The encoder $E_w$ and decoder $D_w$ are trained jointly by minimizing a combined loss, which is defined as:
\begin{equation}
\label{eq:vidmark loss}
    \mathcal{L}_{{vid}}
    = \mathrm{BCE}(\hat{w}, w) + \lambda_0 \,\mathrm{MSE}(v_w, v),
\end{equation}
Here, $\lambda_0$ is a hyperparameter that balances the binary cross entropy (BCE) loss against the mean squared error (MSE) loss, which are used to measure the watermark extraction accuracy and quantify the visual fidelity of watermarked videos, respectively. After training, the decoder $D_w$ is frozen and employed to guide the subsequent fine-tuning of the VGMs.

\subsection{Watermarking VGMs}
\label{sec2_2_watermarking_VGMs}

\begin{figure}[tbp]
    \includegraphics[width=\linewidth]{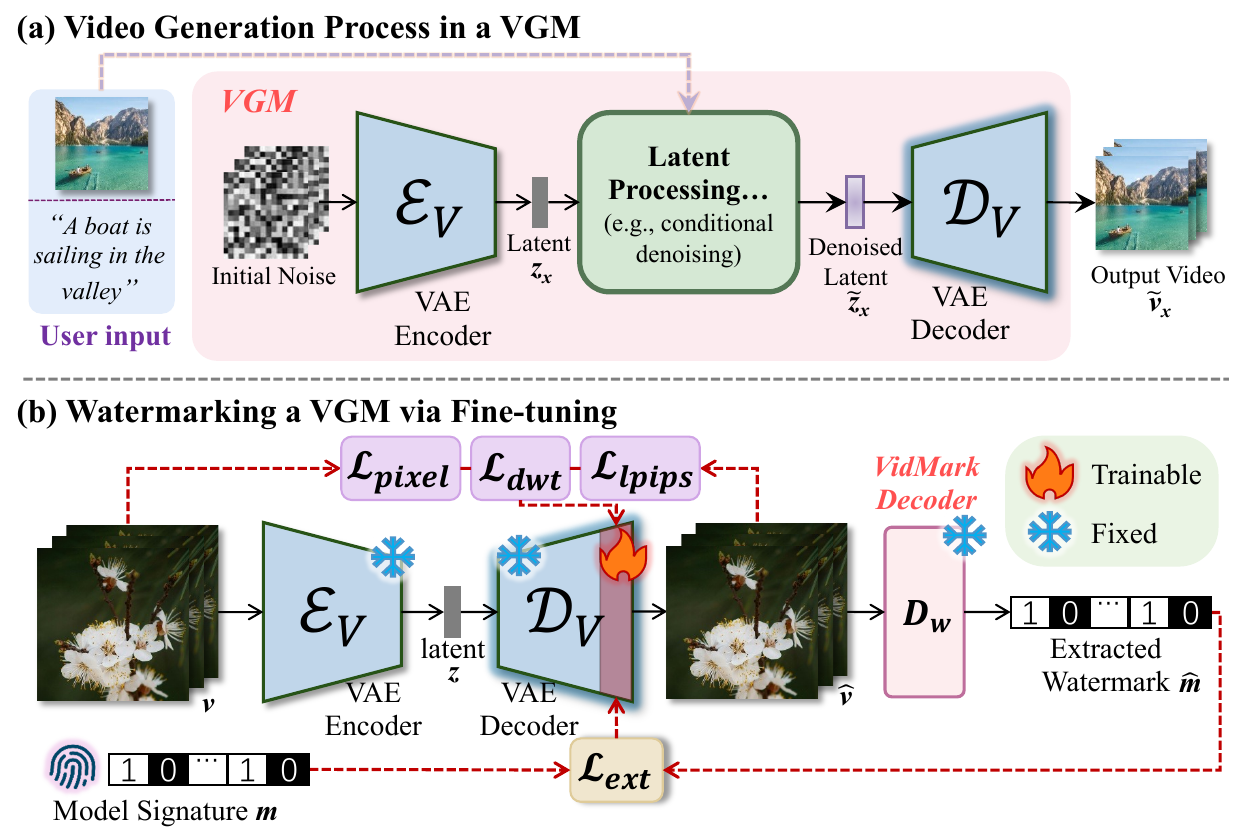}
    \caption{ (a) Illustration of the video generation process in a VGM; (b) Pipeline of VGM watermarking via fine-tuning exploiting the VidMark decoder $D_w$. 
    }
\label{Fig_VGM_pipline}
\end{figure}

This section introduces the generation process in a VGM and describes how to fine-tune VGMs to carry watermarks.

\subsubsection{Video generation process}
Modern VGMs\cite{blattmann2023stable,karras2023dreampose,polyak2024movie,zheng2024open,wan2.1} typically perform video synthesis within a compressed latent space and leverage a video autoencoder (often instantiated as a VAE~\cite{kingma2013auto}) to enable the bidirectional mapping between the pixel and latent spaces~\cite{rombach2022high,blattmann2023align}. This architecture consists of an encoder~$\mathcal{E}_\mathcal{V}$ that maps videos to latents, and a decoder~$\mathcal{D}_\mathcal{V}$ that reconstructs videos from latents. The video generation process is initiated by user inputs, such as text prompts or reference images, which serve as conditions to guide the synthesis. Specifically, as shown in Fig.~\ref{Fig_VGM_pipline}(a), this process starts by sampling an initial random noise vector from a Gaussian distribution within the latent space. A conditional denoising network (e.g., a U-Net or Transformer) then iteratively refines $z_x$ into a denoised latent~$\widetilde{z}_x$ under the guidance of the user inputs. Finally, the VAE decoder $\mathcal{D}_\mathcal{V}$ reconstructs the denoised latent~$\widetilde{z}_x$ back into the pixel space to produce the output video~$\widetilde{v}_x$. Although backbone architectures (e.g., 3D U-Net vs. Diffusion Transformers) and specific VAE implementations may vary, this latent-based generation paradigm remains consistent across modern VGMs \cite{blattmann2023stable,karras2023dreampose,polyak2024movie,zheng2024open,wan2.1}, all of which incorporate a VAE module.

\subsubsection{Fine-tuning pipeline} 
To watermark VGMs, we propose modifying the VAE decoder $\mathcal{D}_\mathcal{V}$ to embed watermarks during the process of reconstructing the video~$\widetilde{v}_x$ from the latent~$\widetilde{z}_x$, ensuring that the final video~$\widetilde{v}_x$ can carry model-specific signatures. This design choice is based on the following considerations. First, $\mathcal{D}_\mathcal{V}$ is responsible for the mapping from latent to pixel space, and, compared to other modules in a VGM, its adjustment requires less computational resources. Second, it is located at the end of the video generation process, and hence the embedded watermark will not be affected by preceding modules (e.g., U-Nets, Transformers). Third, since $\mathcal{D}_\mathcal{V}$ is responsible for a necessary operation in the video generation process, attackers cannot easily bypass or replace the mapping from the latent to the pixel space. Note that, to effectively control the computational overhead, we only selectively fine-tune the $n_f$ layers closest to the output in $\mathcal{D}_\mathcal{V}$.

The fine-tuning pipeline is illustrated in Fig.~\ref{Fig_VGM_pipline}(b). Given an original video~$v \in \mathbb{R}^{F \times 3 \times H \times W}$, the encoder $\mathcal{E}_\mathcal{V}$ maps the video $v$ to a latent representation~$z$, and then the decoder~$\mathcal{D}_\mathcal{V}$ reconstructs it back into pixel space, i.e.,
\begin{equation}
    \hat{v} = \mathcal{D}_V(z),
\end{equation}
where $\hat{v} \in \mathbb{R}^{F \times 3 \times H \times W}$. During the VAE decoder fine-tuning, we use the pre-trained VidMark decoder~$D_w$ as the watermark extractor to predict the watermark sequence~$\hat{m}$ from the output video~$\hat{v}$. Our goal is to enable the VAE decoder~$\mathcal{D}_\mathcal{V}$ to perform the watermark embedding task during the process of reconstructing latent representations into videos, ensuring that each reconstructed video $\hat{v}$ carries a specific model watermark~$m$. Therefore, our method is designed to achieve two primary objectives: (i) minimize the difference between the extracted watermark~$\hat{m}$ and the true model watermark~$m$; and (ii) minimize the distortion between the reconstructed video~$\hat{v}$ and original video~$v$, so as to preserve the consistency of VAE’s bidirectional mapping and avoid interfering with the VGM’s original generation process. 
For objective~(i), we define a watermark extraction loss term $\mathcal{L}_{{ext}}$ as 
\begin{equation}
    \mathcal{L}_{{ext}} = \mathrm{BCE}(\hat{m}, m).
\end{equation}
For objective~(ii), three video fidelity loss terms are introduced to enforce that the reconstructed video $\hat{v}$ matches the original video $v$ in the pixel (spatiotemporal domain), the wavelet domain, and the perceptual feature domain:
\begin{equation}
    \mathcal{L}_{{pixel}} = \mathrm{MSE}(\hat{v}, v),
\end{equation}
\begin{equation}
    \mathcal{L}_{{dwt}} = \mathrm{MSE}\bigl(\mathcal{W}_{\mathcal{LL}}(\hat{v}), \mathcal{W}_{\mathcal{LL}}(v)\bigr),
\end{equation}
\begin{equation}
    \mathcal{L}_{{lpips}} = \frac{1}{N}\sum_{i=1}^{N} \mathrm{LPIPS}\bigl(\hat{v}(i), v(i)\bigr),
\end{equation}
where $\mathcal{W}_{\mathcal{LL}}(\ast)$ indicates the LL subband obtained\ after applying a DWT, and $\mathrm{LPIPS}\bigl(\cdot, \cdot)$ refers to the LPIPS loss~\cite{zhang2018unreasonable} computed between corresponding frames of the reconstructed and that of the original videos.

Overall, a compound loss, i.e.,
\begin{equation}
    \label{eq_loss_finetune}
    \mathcal{L}_{{fine-tune}} = \mathcal{L}_{{ext}} + \lambda_1 \mathcal{L}_{{pixel}} + \lambda_2 \mathcal{L}_{{dwt}} + \lambda_3 \mathcal{L}_{{lpips}},
\end{equation}
is minimized to fine-tune $\mathcal{D}_\mathcal{V}$’s last $n_f$ layers. $\lambda_1$, $\lambda_2$, and $\lambda_3$ are hyperparameters to balance the loss terms.
\section{Watermark Verification Frameworks}
\label{sec3_watermark_verification_framework}


Under the proposed in-generation watermarking scheme, any video produced by a watermarked VGM carries a model-specific watermark to serve as the model signature. This embedding mechanism provides a unified foundation for both synthetic video verification and model ownership verification. To establish a rigorous forensic framework, this section formulates these verification tasks as statistical hypothesis tests. Specifically, we derive theoretical error probabilities and put a constraint on the false positive probability to optimally set the parameters of our framework.

\subsection{Statistical Hypothesis Testing Formulation}

Once a model owner deploys a protected VGM~$\mathcal{V}$ embedded with a $k$-bit model signature~$m$, we define two verification scenarios distinguished by the type of video evidence available.

\subsubsection{Single-Video Scenario}
Given an unknown video $v$, we aim to determine whether it was generated by the protected model $\mathcal{V}$. This is formulated as a video-level hypothesis test, as detailed below.
\begin{itemize}
    \item Null Hypothesis ($\mathcal{H}_0^{(S)}$): The video~$v$ is generated by a non-watermarked model or a model with a different signature.
    \item Alternative Hypothesis ($\mathcal{H}_1^{(S)}$): The video~$v$ is generated by the watermarked model~$\mathcal{V}$.
\end{itemize}
We extract the watermark~$\hat{m}$ from $v$ and compute the Hamming distance~$D(\hat{m}, m)$. The decision rule is: accept~$\mathcal{H}_1^{(S)}$ if $D(\hat{m}, m) \le \tau_1$, where $\tau_1$ is a prescribed threshold; otherwise, accept~$\mathcal{H}_0^{(S)}$.

\subsubsection{Multi-Video Scenario}
Given a set of $n$ videos extracted from $\hat{\mathcal{V}}$,
we aim to verify whether $\hat{\mathcal{V}}$ is a copy of the protected model~$\mathcal{V}$. This is formulated as a model-level hypothesis test, as detailed below.
\begin{itemize}
    \item Null Hypothesis ($\mathcal{H}_0^{(\mathcal{M})}$): The suspect VGM~$\hat{\mathcal{V}}$ is independent of $\mathcal{V}$ (e.g., $\hat{\mathcal{V}}$ is a non-watermarked model or a model with a different signature).
    \item Alternative Hypothesis ($\mathcal{H}_1^{(\mathcal{M})}$): The suspect VGM~$\hat{\mathcal{V}}$ is a copy of $\mathcal{V}$.
\end{itemize}
We extract the watermark from each video and concatenate them to form an aggregate sequence $\hat{M}$ of length $n \cdot k$, where $k$ is the length of signature embedded in the VGM. This sequence is compared against the true signature sequence $M$, which is concatenated from $n$~copies of the model signature~$m$. The decision rule is: accept $\mathcal{H}_1^{(\mathcal{M})}$ if  $D(\hat{M}, M) \le \tau_n$, where $\tau_n$ is a prescribed threshold; otherwise, accept $\mathcal{H}_0^{(\mathcal{M})}$.

\begin{figure}[tbp]
    \includegraphics[width=\linewidth]{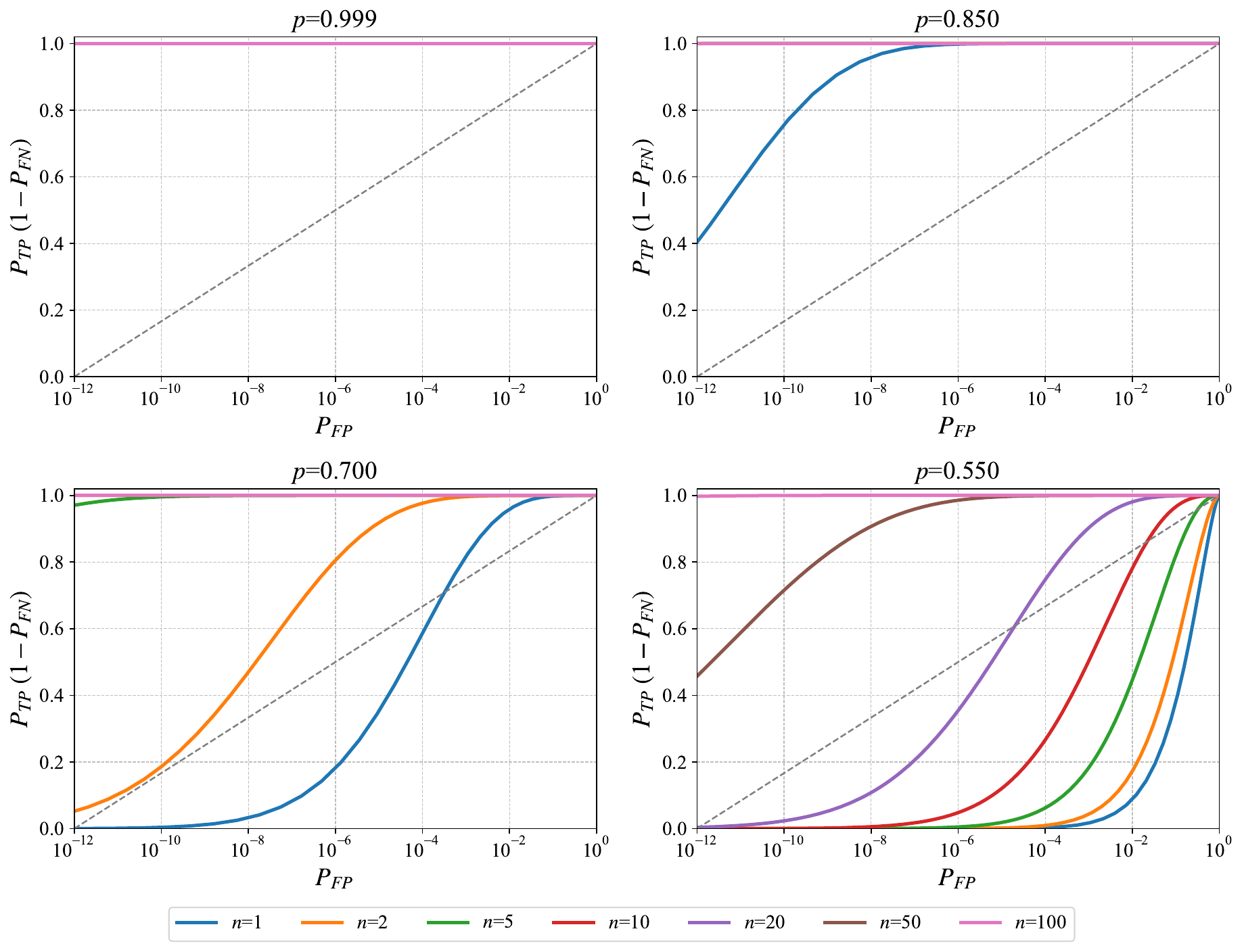}
    \centering
    \caption{ Theoretical verification performance in multi-video scenarios. The figure presents ROC curves, showing the values of $P_{FP}$ and $P_{TP}$ at different bitwise extraction accuracy ($p$) and video counts ($n$). In the subfigures for $p$ = 0.999, 0.850, and 0.700, certain underlying curves are entirely occluded by the topmost curve ($n$ = 100). Note that, for simplicity, only the results for $k=96$ are shown here; other $k$ values show similar evolution patterns. 
    }
\label{Fig_ROC}
\end{figure}

\subsection{Theoretical Probability Analysis}
To quantify the reliability of our frameworks, we analyze the theoretical error probabilities. Our derivation  focuses on the multi-video scenario, as the single-video scenario is mathematically equivalent to the special case of the multi-video scenario with $n=1$. Furthermore, regarding the null hypothesis $\mathcal{H}_0^{(\mathcal{M})}$, we only consider the case of a non-watermarked model, 
which is statistically indistinguishable from a model embedded with a different signature~\cite{fernandez2023stable}.

\vspace{3pt}
\noindent \textbf{False Positive Probability.}
A false positive event occurs when a non-watermarked model accidentally produces videos whose extracted signature matches the true model’s signature. Assuming that the extracted bits are independent and uniformly distributed, the number of matching bits $X$ follows a Binomial distribution $X \sim \mathcal{B}(nk, 0.5)$. Since 
$D(\hat{M}, M) = nk - X$, $P_{FP}$ is the probability that $nk - X \le \tau_n$. If $nk$ is large, according to the Central Limit Theorem, the false positive probability is:
\begin{equation}
\label{eq_p_fp}
    P_{FP} = \sum_{x=nk-\tau_n}^{nk} \binom{nk}{x} 0.5^{nk} \approx 1 - \Phi\left(\frac{nk - 2\tau_n}{\sqrt{nk}}\right),
\end{equation}
where $\Phi(\cdot)$ denotes the cumulative distribution function (CDF) of the standard normal distribution. 

\vspace{3pt}
\noindent \textbf{False Negative Probability.}
A false negative event occurs when, in the case of a watermarked model, the extracted signature fails to match the target signature due to attacks or distortions, resulting in  $D(\hat{M}, M) > \tau_n$. Given a bitwise extraction accuracy $p$, assuming that bit errors are independent of each other, the number of matching bits $X$ follows a binomial distribution $X \sim \mathcal{B}(nk, p)$. If $nk$ is large, by the Central Limit Theorem, 
the false negative probability is:
\begin{equation}
\label{eq_p_fn}
    P_{FN} = \sum_{x=0}^{nk-\tau_n-1} \binom{nk}{x} p^x (1-p)^{nk-x} \approx \Phi\left(\frac{nk - \tau_n - nkp}{\sqrt{nkp(1-p)}}\right).
\end{equation}
where $\Phi(\cdot)$ denotes the CDF of the standard normal distribution as above. 

It is worth noting that some existing studies (e.g., \cite{fernandez2023stable}) have observed that the watermark bits extracted from non-watermarked generative models may exhibit slight zero-one biases at certain bit positions, thereby deviating from a strictly independent and identically distributed (i.i.d.) Bernoulli(0.5) assumption. Although this deviation may slightly affect the precise theoretical error probability, the mathematical formulas 
(Eqs.~(\ref{eq_p_fp}) and (\ref{eq_p_fn}))
can still be used to select the key parameters such as decision thresholds and video counts (see Sections~\ref{sec: Theoretical Performance and Decision Strategy} and~\ref{sec: VGM Experiments Results} for details). Furthermore, the empirical values, i.e., true positive rate (TPR) and false positive rate (FPR), calculated through extensive experiments in Section~\ref{sec: VGM Experiments Results} demonstrate that the deviation mentioned in \cite{fernandez2023stable} will not affect the reliability of our verification frameworks.

\subsection{Theoretical Performance and Decision Strategy}
\label{sec: Theoretical Performance and Decision Strategy}
Unlike previous watermarking schemes that rely on single-sample detection~\cite{fernandez2023stable,kim2024wouaf,liu2025implanting}, our verification framework can utilize the law of large numbers, using $n$ generated videos to accumulate evidence to achieve model ownership verification. 
We evaluated the proposed model ownership verification framework by considering different pairs of $p$ and $n$, and varying $\tau_n$ from $0$ to $nk$, thereby obtaining a sequence of operating points ($P_{FP}$, $1-P_{FN}$) 
yielding  the ROC curve. As shown in Fig.~\ref{Fig_ROC}, as $n$ increases, the ROC curves shift significantly to the upper-left corner, indicating a significant improvement in theoretical verification performance. Notably, even when the bitwise extraction accuracy $p$ drops significantly to near-random guessing levels (e.g., $p=0.550$), accumulating evidence from multiple videos (e.g., $n\ \geq20$) rapidly shifts the ROC curve to the upper-left corner. 
This confirms that 
reliable model ownership verification  can be guaranteed 
by simply increasing the number of videos $n$.
In the field of digital forensics, avoiding false accusations is the primary concern~\cite{schaathun2006boneh,barni2021dnn}. Therefore, based on the established theoretical derivation, we put a constraint on the $P_{FP}$ to optimally set the parameters
of our framework.
First, the threshold $\tau_n$ is strictly selected to ensure that $P_{FP}\le{10}^{-6}$. Then, we minimize the theoretical $P_{FN}$ by selecting an appropriate number of queries $n$. The detailed parameter configuration of the verification frameworks (i.e., $k$, $\tau_1$, $\tau_n$, $n$) and their corresponding $P_{FP}$ and $P_{FN}$ will be provided in Section~\ref{sec: VGM Experiments Results}.

\section{Evaluation of VidMark on Video Watermarking}
\label{sec4_experiments_on_video_watermarking}
Before integrating VidMark into the VGM fine-tuning pipeline, this section evaluates its standalone performance on the traditional video watermarking task. Specifically, Section~\ref{sec4_1_experimental_settings_of_VidMark} details the experimental setup, encompassing the dataset, training configurations, evaluation metrics, and baselines. Subsequently, Section~\ref{sec4_2_experimental_results_of_VidMark} provides a comprehensive comparative analysis against state-of-the-art baselines, demonstrating VidMark's superiority in balancing visual fidelity and robust watermark extraction. Finally, Section~\ref{sec4_3_ablation_studies_of_VidMark} presents ablation studies to validate the effectiveness of the specially designed components in VidMark, i.e., the DWT and GTAB.

\subsection{Experimental Settings}
\label{sec4_1_experimental_settings_of_VidMark}

\subsubsection{Implementation details}
We construct our dataset from Pexels Videos\footnote{\hyperlink{https://www.pexels.com}{https://www.pexels.com}}, consisting of 40k video clips for training and 10k for evaluation. Each clip contains 16 consecutive frames, cropped to a resolution of 512$\times$512. VidMark is trained for 10 epochs (25,000 iterations in total) on a single NVIDIA A100 (80 GB) GPU with a batch size of 16, and optimized using the Adam optimizer~\cite{kingma2014adam} with a learning rate of ${10}^{-4}$ to minimize Eq. (\ref{eq:vidmark loss}). We adopt a dynamic scheduling strategy for the hyperparameter $\lambda_0$: it is initialized to 0 for the first 2,500 iterations, then increased to 1, and finally set to 10 after 15,000 iterations. During training, a random $k$-bit watermark is generated for each input video.

\subsubsection{Evaluation Metrics}
To evaluate the watermark extraction accuracy, we use bitwise accuracy (Bit Acc) to measure the proportion of matching bits between the ground truth watermark and the extracted watermark. To assess visual quality, we include the Learned Perceptual Image Patch Similarity (LPIPS)~\cite{zhang2018unreasonable} to quantify the average difference between the frames in watermarked videos and non-watermarked videos. Crucially, to capture the spatiotemporal consistency and distribution shifts of the watermarked videos, Fr\'{e}chet Video Distance (FVD)~\cite{Unterthiner2019FVDAN} and ColorVideoVDP (CVVDP)~\cite{mantiuk2024colorvideovdp} are employed as two additional video quality metrics.

\begin{table*}[ht]
\centering
\scriptsize   
\setlength{\tabcolsep}{0.5mm}{}
\renewcommand\arraystretch{1}
\caption{
    Comparison of different methods for video watermarking on Pexels-512 dataset.
}
\begin{tabular}{lcccccccccccccccccc}
\toprule
\multirow{4}{*}{\textbf{Method}} & \multirow{4}{*}{\shortstack{\textbf{Payload}\\(bits)}} & \multicolumn{3}{c}{\textbf{Video Fidelity}} & \multicolumn{14}{c}{\textbf{Bit Acc} $\uparrow$} \\
\cmidrule(lr){3-5}
\cmidrule(lr){6-19}

 &  & \multirow{3}{*}{\shortstack{FVD\\$\downarrow$}} & \multirow{3}{*}{\shortstack{CVVDP\\$\uparrow$}} & \multirow{3}{*}{\shortstack{LPIPS\\$\downarrow$}} & \multirow{3}{*}{\shortstack{No\\Attack}} & Motion & H.264/ & Frame & Frame & Temporal & Gaussian & Gaussian & \multirow{2}{*}{AV1} & Scale & Brightness & Contrast & Overlay & Overlay \\
 &  &  &  &  &  & Blur & AVC & Average & Drop & Jitter & Noise & Blur &  & Adjust & Adjust & Adjust & Text & Emoji \\
 &  &  &  &  &  & std=5 & CRF=27 & $N$=4 & $p$=0.5 & $N$=3 & $\sigma$=0.02 & std=5 & CRF=32 & $p$=0.75 & $f$=0.5 & $f$=0.5 & 24$\times$8px & 25$\times$25px \\
\midrule
Dct-Dwt~\cite{navas2008dwt} & 16 & 82.85 & 8.48 & 0.248 & 0.77 & 0.54 & 0.52 & 0.69 & 0.79 & 0.77 & 0.81 & 0.54 & 0.50 & 0.68 & 0.48 & 0.63 & 0.77 & 0.77 \\
HiDDeN~\cite{zhu2018hidden} & 48 & 57.72 & 8.82 & 0.362 &\textbf{1.00}& 0.57 & 0.57 & 0.65 & 0.77 & 0.78 & 0.77 & 0.57 & 0.61 & 0.66 & 0.91 & 0.80 & 0.78 & 0.78 \\
CIN~\cite{ma2022towards} & 30 & 27.47 & 8.65 & 0.111 &\textbf{1.00}&\textbf{1.00}& 0.80 &\textbf{1.00}& 0.57 &\textbf{1.00}&\textbf{1.00}&\textbf{1.00}& 0.81 &\textbf{1.00}&\textbf{1.00}&\textbf{1.00}&\textbf{1.00}&\textbf{1.00}\\
TrustMark~\cite{bui2023trustmark} & 100 & 6.58 & 9.32 & 0.046 &\textbf{1.00}& 0.61 & 0.58 & 0.62 & 0.61 & 0.61 & 0.61 & 0.61 & 0.58 & 0.61 & 0.66 & 0.67 & 0.61 & 0.61 \\
REVMark~\cite{zhang2023novel} & 96 & 37.01 & 7.92 & 0.127 &\textbf{1.00}&\textbf{1.00}&\textbf{1.00}&\textbf{1.00}& 0.99 & 0.97 &\textbf{1.00}&\textbf{1.00}& 0.97 &\textbf{1.00}&\textbf{1.00}&\textbf{1.00}&\textbf{1.00}&\textbf{1.00}\\
\rowcolor{gray!8} & 32 & \textbf{5.07} & \textbf{9.36} & \textbf{0.027} &\textbf{1.00}&\textbf{1.00}& 0.99 &\textbf{1.00}&\textbf{1.00}&\textbf{1.00}&\textbf{1.00}&\textbf{1.00}& 0.98 &\textbf{1.00}&\textbf{1.00}&\textbf{1.00}&\textbf{1.00}&\textbf{1.00}\\
\rowcolor{gray!8} & 48 & 7.87 & 9.32 & 0.030 &\textbf{1.00}&\textbf{1.00}&\textbf{1.00}&\textbf{1.00}&\textbf{1.00}&\textbf{1.00}&\textbf{1.00}&\textbf{1.00}& 0.98 &\textbf{1.00}&\textbf{1.00}&\textbf{1.00}&\textbf{1.00}&\textbf{1.00}\\
\rowcolor{gray!8} & 64 & 6.64 & 9.28 & 0.032 &\textbf{1.00}&\textbf{1.00}& 0.98 &\textbf{1.00}&\textbf{1.00}&\textbf{1.00}&\textbf{1.00}&\textbf{1.00}& 0.97 &\textbf{1.00}&\textbf{1.00}&\textbf{1.00}&\textbf{1.00}&\textbf{1.00}\\
\rowcolor{gray!8} & 96 & 6.74 & 9.23 & 0.030 &\textbf{1.00}&\textbf{1.00}& 0.99 &\textbf{1.00}& 0.99 &\textbf{1.00}&\textbf{1.00}&\textbf{1.00}& 0.96 &\textbf{1.00}&\textbf{1.00}&\textbf{1.00}&\textbf{1.00}&\textbf{1.00}\\
\rowcolor{gray!8} \multirow{-5}{*}{\shortstack{VidMark\\(\textbf{\textit{Ours}})}} & 128 & 31.18 & 8.60 & 0.043 &\textbf{1.00}&\textbf{1.00}& 0.99 &\textbf{1.00}& 0.99 &\textbf{1.00}&\textbf{1.00}&\textbf{1.00}& \textbf{0.99} &\textbf{1.00}&\textbf{1.00}&\textbf{1.00}&\textbf{1.00}& \textbf{1.00}\\
\bottomrule
\end{tabular}

\vspace{1.5mm}

\begin{minipage}{\linewidth}
\scriptsize
\textit{Note:} 
The video-level distortions and their corresponding parameters are defined as follows. \textbf{Motion Blur and Gaussian Blur}: video filtering with a kernel standard deviation of $std$; \textbf{H.264/AVC and AV1}: video compression utilizing a Constant Rate Factor (CRF); \textbf{Frame Average}: temporal averaging over $N$ consecutive frames; \textbf{Frame Drop}: randomly dropping frames with a ratio of $p$, where dropped frames are padded with zeros to preserve temporal length; \textbf{Temporal Jitter}: randomly swapping frames within a local temporal window of size $N$; \textbf{Gaussian Noise}: applying additive zero-mean Gaussian noise with a variance of $\sigma$; \textbf{Scale Adjust}: spatial downsampling by a ratio of $p$ followed by upsampling to the original resolution; \textbf{Brightness Adjust and Contrast Adjust}: linearly scaling the brightness or contrast levels by a factor of $f$; \textbf{Overlay Text and Overlay Emoji}: superimposing a specific patch onto the frames.
\end{minipage}

\label{tab:VidMark performance}

\end{table*}

\subsubsection{Comparison Methods}
To comprehensively evaluate VidMark’s performance on video watermarking, we compare it against five representative watermarking schemes spanning three categories: traditional methods, deep image watermarking methods, and deep video watermarking methods. These baselines include:  

(i) Traditional method: Dct-Dwt~\cite{navas2008dwt}, a classic hybrid frequency-domain approach that serves as a baseline for non-learning techniques.  

(ii) Deep image watermarking methods: HiDDeN~\cite{zhu2018hidden}, CIN~\cite{ma2022towards}, and TrustMark~\cite{bui2023trustmark}. HiDDeN is a pioneering encoder-decoder architecture; CIN emphasizes invertibility for high fidelity; and TrustMark focuses on restoration and robustness. Since these models are designed for static images, we apply them to video data in a frame-wise manner.

(iii) Deep video watermarking method: REVMark~\cite{zhang2023novel}, a recent state-of-the-art video-specific watermarking framework. To ensure a fair comparison, all baseline methods are either retrained or evaluated on the constructed Pexels Videos dataset using their official open-source implementations under identical settings. Note that some baseline methods, specifically CIN~\cite{ma2022towards}, TrustMark~\cite{bui2023trustmark}, and REVMark~\cite{zhang2023novel}, encounter difficulties when directly processing high-resolution video data (512$\times$512). The official pre-trained checkpoints of these methods are typically optimized for lower resolutions (e.g., 128$\times$128 or 256$\times$256).
For fair and feasible comparison, we adopt a resize-embed-resize strategy for these baselines. Specifically, the original 512$\times$512 videos are downscaled to the resolutions supported by the corresponding pre-trained watermarking models, embedded with watermarks, and then upscaled back to 512$\times$512 for visual quality evaluation. For watermark extraction, the watermarked videos are resized to the resolutions required by the respective extractors.

\subsection{Experimental Results}
\label{sec4_2_experimental_results_of_VidMark}
Table~\ref{tab:VidMark performance} reports quantitative results on the Pexels Videos dataset. VidMark significantly outperforms existing state-of-the-art methods in video fidelity across all metrics. Notably, VidMark with a 32-bit payload achieves the best FVD of 5.07 and LPIPS of 0.027, indicating that the watermarked videos remain visually indistinguishable from the originals. By contrast, REVMark~\cite{zhang2023novel} obtains a much higher FVD of 37.01, suggesting noticeable temporal artifacts. Compared with image-based methods such as HiDDeN~\cite{zhu2018hidden}, VidMark also achieves better video-specific metrics, i.e., FVD and CVVDP, demonstrating its effectiveness in preserving temporal consistency in high-resolution videos.

Regarding robustness, VidMark demonstrates exceptional resilience against diverse attacks while maintaining high embedding capacity. As shown in Table~\ref{tab:VidMark performance}, traditional and deep image watermarking methods (e.g., Dct-Dwt~\cite{navas2008dwt} and HiDDeN~\cite{zhu2018hidden}) struggle with temporal distortions, such as frame averaging and temporal jitter, with bit accuracy dropping to near-random guessing (0.50-0.70). Conversely, VidMark maintains over 0.99 Bit Acc across almost all degradation types, including severe H.264/AV1 compression and combined temporal manipulations. Even if the payload is increased to 128 bits, VidMark can still maintain a high extraction accuracy, significantly outperforming the competing schemes such as TrustMark~\cite{bui2023trustmark}, which cannot accurately extract watermarks under most attacks.
It is worth noting that VidMark achieves such high robustness {\em without introducing any simulated attacks} (e.g., differentiable noise layers) during training. This robustness primarily stems from its architectural design: the VidMark encoder uses a 3D convolutional backbone network and GTAB to distribute watermark information redundantly across the global spatiotemporal dimensions of the video. Therefore, even if local spatial details are compressed or temporal continuity is disrupted (e.g., through frame dropping or temporal jitter), the decoder can still reliably aggregate the remaining information to fully recover the embedded watermark.

\begin{figure}[tbp]
    \centering
    \includegraphics[width=.9\linewidth]{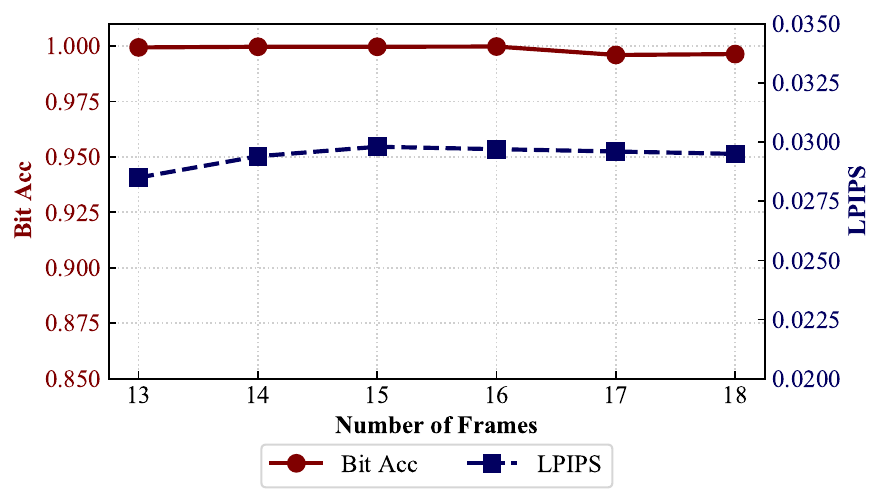}
    \centering
    \caption{ 
    Results under different numbers of frames.
    }
\label{Fig_multiple_video}
\end{figure}

Furthermore, this global spatiotemporal aggregation mechanism endows VidMark with excellent temporal flexibility. As shown in Fig.~\ref{Fig_multiple_video}, VidMark is trained only on 16-frame videos with an embedding payload of 96 bits, yet during the inference process, the videos range from 13 to 18 frames. Under these varying conditions, VidMark still maintains Bit Acc close to 1.00 and low LPIPS. This confirms that the proposed VidMark does not overfit to the fixed temporal dimension during training, but seamlessly generalizes to variable-length video inputs without affecting the watermark embedding/extraction process or perceptual quality of watermarked videos.

\begin{table}[t]
    \centering
    \setlength{\tabcolsep}{3mm}{}
    \renewcommand\arraystretch{1}
    \caption{
            Ablation study of VidMark's various components. 
        }
    \begin{tabular}{ccccc} 
    \toprule
        DWT &  GTAB & CVVDP $\uparrow$ & LPIPS $\downarrow$ & Bit Acc $\uparrow$\\
    \midrule
        \ding{55} & \ding{55} & 8.82  & 0.039 & 0.96 \\
        \ding{51} & \ding{55} & 8.87 & 0.038 & 1.00 \\
        \rowcolor{gray!8} 
        \ding{51} & \ding{51} & \textbf{9.23} & \textbf{0.030}  & \textbf{1.00} \\
    \bottomrule
 \end{tabular}
\label{tab:ablation study on DWT and STAB}
\end{table}

\subsection{Ablation Studies}
\label{sec4_3_ablation_studies_of_VidMark}
To investigate the contribution of the core components in VidMark, i.e., the two-scale DWT decomposition and the GTAB, an ablation study is conducted. The quantitative results are summarized in Table~\ref{tab:ablation study on DWT and STAB}. We first evaluate the baseline model without DWT and GTAB (the first row in Table~\ref{tab:ablation study on DWT and STAB}), where the watermark is embedded directly in the spatial domain. As shown in the second row, introducing the two-scale DWT decomposition significantly boosts the Bit Acc from 0.96 to 1.00, while also improving the CVVDP score from 8.82 to 8.87. This validates that embedding watermarks into the low-frequency subband of the wavelet domain provides a more robust representation than the spatial domain. 

Then, we further incorporate the GTAB. Comparing the second and third rows, while the Bit Acc remains at 1.00, the visual fidelity metrics show substantial improvement. Specifically, LPIPS decreases significantly from 0.039 to 0.030, and CVVDP further increases to 9.23. This indicates that the GTAB effectively captures complex spatiotemporal dependencies across frames, enabling the model to better conceal watermark artifacts and generate high-fidelity watermarked videos. Consequently, the full VidMark model (the third row) achieves the best trade-off between robustness and imperceptibility.
\section{Experiments on VGM Watermarking}
\label{sec5_experiments_on_VGM_watermarking}
This section comprehensively evaluates the effectiveness and generalization ability of the proposed in-generation watermarking scheme. By leveraging the pre-trained VidMark decoder, we fine-tune three representative state-of-the-art VGMs, including Wan~\cite{wan2.1}, Open-Sora~\cite{zheng2024open}, and Stable Video Diffusion (SVD)~\cite{blattmann2023stable}. The implementation details are first described in Section~\ref{sec5_1_experimental_settings_of_VGM_watermarking}. Subsequently, Section~\ref{sec: VGM Evaluation Metrics} outlines the evaluation metrics used in our experiments. Finally, Section~\ref{sec: VGM Experiments Results} presents a comprehensive analysis of the experimental results, including the impact on video generation, watermark robustness, computational complexity, and the VidMark decoder’s specific contribution.

\begin{table}[t]
\centering
\caption{Theoretical error probabilities under the selected parameters.
}
\label{tab:theoretical_bounds}
\resizebox{\columnwidth}{!}
{
\begin{tabular}{lccccccc}
\toprule
\multirow{2}{*}{\textbf{Scenario}} & \multirow{2}{*}{\textbf{VGM}} &  \multirow{2}{*}{$k$} &  \multirow{2}{*}{$n$} & \multirow{2}{*}{$\tau$} & \multirow{2}{*}{$P_{FP}$} & $P_{FN}$ & $P_{FN}$  \\
& & & & & &  ($p=0.95$)  & ($p=0.75$)\\
\midrule
\multirow{2}{*}{\shortstack{Single-\\video}} & SVD & 64 & 1 & 13 & $9.41 \times 10^{-7}$ & $2.72 \times 10^{-6}$ & $76.10\%$ \\
 & Open-Sora / Wan & 96 & 1 & 24 & $4.85 \times 10^{-7}$ & $< 10^{-10}$ & $44.54\%$ \\
\midrule
\multirow{2}{*}{\shortstack{Multiple-\\video}} & SVD & 64 & 20 & 554 & $8.49 \times 10^{-7}$ & $< 10^{-10}$ & $< 10^{-10}$ \\
 & Open-Sora / Wan & 96 & 20 & 855 & $9.02 \times 10^{-7}$ & $< 10^{-10}$ & $< 10^{-10}$ \\
\bottomrule
\end{tabular}
}
\end{table}

\subsection{Implementation Details}
\label{sec5_1_experimental_settings_of_VGM_watermarking}

Contemporary VGMs are primarily based on two generation paradigms: text-to-video (T2V) and image-to-video (I2V) synthesis. We apply our scheme across three state-of-the-art, fully open-source VGMs, i.e., Wan~\cite{wan2.1}, Open-Sora~\cite{zheng2024open}, and Stable Video Diffusion (SVD)~\cite{blattmann2023stable}. Specifically, in our experiments, Wan~\cite{wan2.1} and Open-Sora~\cite{zheng2024open} are employed for T2V synthesis, while SVD~\cite{blattmann2023stable} is configured for I2V synthesis.

To embed the watermark, we exclusively optimize the learnable parameters within the last few layers of the VAE decoder $\mathcal{D}_\mathcal{V}$, while freezing all other modules and layers, including the VAE encoder, the generative backbones (e.g., DiT or U-Net), and the preceding layers of the VAE decoder. This simplification strategy significantly reduces computational overhead, allowing the entire fine-tuning process to be performed on a single NVIDIA A100 (80 GB) GPU. During the fine-tuning phase, a $k$-bit watermark is randomly generated and subsequently fixed as the target model signature $m\in\{0,1\}^k$.

In the forward propagation, the frozen VAE encoder projects the input videos into the latent representations, which are then mapped back to the pixel space by the VAE decoder. The reconstructed videos are subsequently fed into the pre-trained VidMark decoder to extract the embedded watermark $\hat{m}$. In the backward propagation, by minimizing the composite loss in Eq. (\ref{eq_loss_finetune}), the learnable parameters within the last few layers of the VAE decoder are iteratively updated.

For this fine-tuning process, we utilize a video subset collected from Pexels Videos, comprising 40k videos for training and 10k videos for evaluation. To accommodate the high-resolution synthesis capabilities and diverse temporal modeling mechanisms across different VGMs, the length of the input video ranges from 5 to 16 frames, with a resolution of $512\times512$ per frame. The learnable parameters are optimized using the Adam optimizer~\cite{kingma2014adam} with a learning rate of $5\times{10}^{-4}$, and the training takes over 3 epochs with a batch size of 1.

\begin{table*}[t]
    \centering
    \setlength{\tabcolsep}{4.0mm}{}
    \renewcommand\arraystretch{1}
    \caption{Comparison of different methods on video quality and watermark extraction. 
    }
    \begin{tabular}{l|c|cccccc|c} 
    \toprule
        \multirow{2}{*}{\textbf{Method}} & \multirow{2}{*}{\shortstack{\textbf{Payload}\\(bits)}} & \multicolumn{6}{c|}{\textbf{Video Quality}} & \multirow{2}{*}{\textbf{Bit Acc}}\\
        & & VMAF $\uparrow$  & CVVDP $\uparrow$  & PSNR $\uparrow$ & LPIPS $\downarrow$ & SSIM $\uparrow$ & FVD $\downarrow$ & \\
    \midrule

    \multicolumn{9}{c}{Watermarking Wan model~\cite{wan2.1}}\\
        \midrule
        \textit{Non-watermarked} & 0 & \textit{95.72} & \textit{9.795}  & \textit{41.81} & \textit{0.008} & \textit{0.974}  & \textit{0.34} & \textit{0.48}\\
        \multirow{1}{*}{Stable Sig.~\cite{fernandez2023stable}} & 48 & 72.57 & 7.075 & 30.82 & 0.284  & 0.783 & 191.07 & \textbf{1.00}\\
        WOUAF~\cite{kim2024wouaf} & 32  & 91.05 & 7.289 & 28.92 & 0.073  & \textbf{0.904} & 77.82 & 0.72\\
        \rowcolor{gray!8}\multirow{-1}{*}{\shortstack{\textbf{\textit{Our Scheme}}}} & \textbf{96}  & \textbf{92.36}  & \textbf{9.086}  & \textbf{34.61}  & \textbf{0.059} & 0.881  & \textbf{15.57} & \textbf{1.00}\\
        \midrule
        
    \multicolumn{9}{c}{Watermarking Open-Sora model \cite{zheng2024open}}\\
        \midrule
        \textit{Non-watermarked} & 0 & \textit{92.86} & \textit{9.201} & \textit{39.28}  & \textit{0.042}  &  \textit{0.947} & \textit{4.90} & \textit{0.55} \\
        \multirow{1}{*}{Stable Sig.~\cite{fernandez2023stable}} & 48 & 13.19 & -0.508 & 19.88 & 0.682 & 0.367 & 751.50 & \textbf{1.00}\\
        WOUAF~\cite{kim2024wouaf} & 32 & 77.70 & 6.169 & 24.40 &  0.124 & 0.841 & 214.71 & 0.75\\
        \rowcolor{gray!8}\multirow{-1}{*}{\shortstack{\textbf{\textit{Our Scheme}}}} & \textbf{96}  &  \textbf{83.39} & \textbf{7.918}  & \textbf{34.55}  & \textbf{0.112} & \textbf{0.897} &  \textbf{77.72} & \textbf{1.00}\\
        \midrule
    
    \multicolumn{9}{c}{Watermarking SVD model \cite{blattmann2023stable}}\\
        \midrule
        \textit{Non-watermarked} & 0 & \textit{90.20} & \textit{9.281} & \textit{36.99} & \textit{0.027} & \textit{0.939}  & \textit{3.70} & \textit{0.47}\\
        Stable Sig. \cite{fernandez2023stable} & 48 & 83.21 & 8.526  & 34.06 & 0.234 & 0.831 & 28.51 & 1.00\\
        WOUAF~\cite{kim2024wouaf} &  32 & \textbf{93.45} & 8.730 & 30.56 & \textbf{0.046} & \textbf{0.928} & \textbf{10.66} & 0.76\\
        \rowcolor{gray!8} 
        \textbf{\textit{Our Scheme}} & \textbf{64}  & 93.01 & \textbf{9.103} & \textbf{35.20} &   0.050 &  0.890 &  13.13 & \textbf{1.00}\\
        
    \bottomrule
    \end{tabular}
    \label{tab:main table}
\end{table*}

\begin{figure*}[t]
\centering
\includegraphics[width=.95\linewidth]{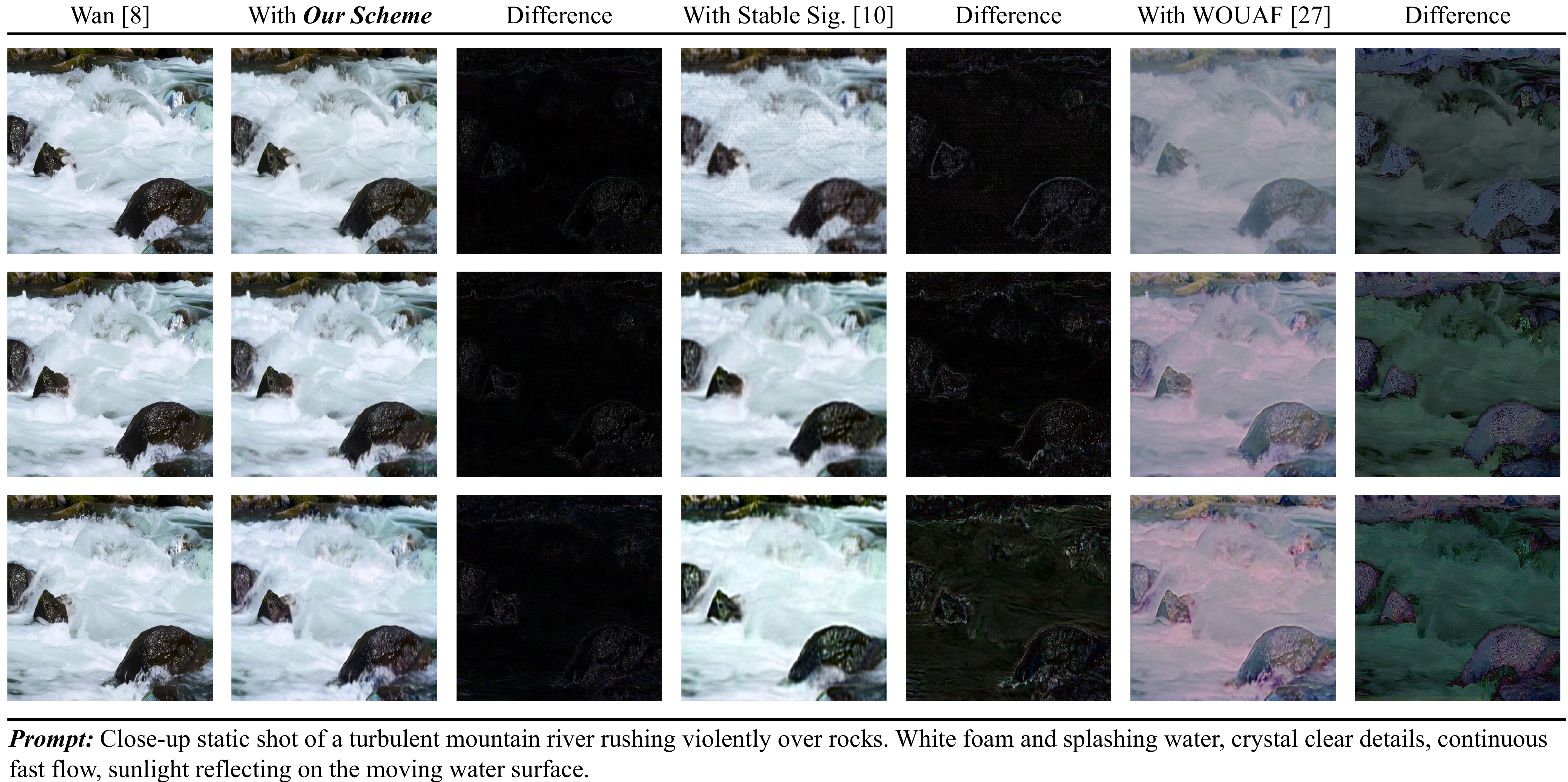}
\caption{
    Visual comparison of videos generated by the original Wan model~\cite{wan2.1} and its watermarked counterparts using different schemes, conditioned on identical text prompts. The pixel  differences between the watermarked videos and original videos are also provided.
}
\label{Fig_visual_comparison}
\end{figure*}

\subsection{Evaluation Metrics}
\label{sec: VGM Evaluation Metrics}
In our experiments, we comprehensively evaluate the performance of the proposed watermarking scheme from three aspects: video fidelity, watermark extraction accuracy, and watermark verification accuracy.

(i) {Video fidelity:} We employ both frame-level metrics (e.g., PSNR, SSIM, and LPIPS~\cite{zhang2018unreasonable}) and video-level spatiotemporal metrics (e.g., Fr\'{e}chet Video Distance (FVD)~\cite{Unterthiner2019FVDAN}, VMAF~\cite{li2016vmaf}, and ColorVideoVDP (CVVDP)~\cite{mantiuk2024colorvideovdp}) to evaluate the video fidelity. Specifically, these metrics are computed by comparing the reconstructed videos (i.e., output of the watermarked or original VAE decoder) with the original videos (i.e., input of the VAE encoder).

(ii) {Watermark extraction accuracy:} The bitwise accuracy (i.e., Bit Acc) between the true model signature and the extracted watermark is employed to measure the watermark extraction performance, which is evaluated under both non-attack and various attack scenarios.

(iii) Watermark verification accuracy: We evaluate the watermark verification accuracy by measuring the True Positive Rate (TPR) and False Positive Rate (FPR) following the verification framework detailed in Section~\ref{sec3_watermark_verification_framework}. 
To meet the forensic-level security requirements, strict thresholds $\tau_1$ and $\tau_n$ are established under the condition that the $P_{FP}$~$<{10}^{-6}$.

Specifically, owing to the varying watermark capacities of different generative architectures, the signature length k is set to 64 for SVD~\cite{blattmann2023stable} and 96 for Open-Sora~\cite{zheng2024open} and Wan~\cite{wan2.1}. The corresponding thresholds are configured as follows: for the single-video scenario ($n=1$), $\tau_1=13$ ($k$=64) and $\tau_1$=24 ($k$=96); for the multiple-video scenario, we set the number of videos $n$ to 20, with $\tau_{20}=554$ (k=64) and $\tau_{20}$=855 ($k$=96). The detailed parameter configurations for single-video and multiple-video scenarios, alongside their $P_{FP}$ and $P_{FN}$, are summarized in Table~\ref{tab:theoretical_bounds}. The empirical values of TPR and FPR are subsequently measured over 10,000 and 500 independent trials for the single-video and multiple-video scenarios, respectively.

\subsection{Experimental Results and Analysis}
\label{sec: VGM Experiments Results}
For comparison, the original, non-watermarked VGMs are used as baselines. Furthermore, since there are currently no open-source in-generation watermarking solutions specifically for VGMs, we incorporate two representative baseline frameworks originally designed for IGMs (i.e., Stable Sig.~\cite{fernandez2023stable} and WOUAF~\cite{kim2024wouaf}) on a frame-by-frame basis.

The experimental results are presented from four aspects. First, the impact of different watermarking schemes on video generation quality is evaluated by measuring video fidelity metrics, including PSNR, SSIM, LPIPS, FVD, VMAF, and CVVDP. Second, the watermark robustness against various video-level and model-level attacks is assessed using Bit Acc, TPR, and FPR. Third, the computational complexity is analyzed in terms of memory cost and inference time. Finally, ablation studies are conducted to verify the effectiveness of the pre-trained VidMark decoder in watermarking VGMs.

\begin{table*}[htbp]
    \centering
    \setlength{\tabcolsep}{0.7mm}
    \renewcommand\arraystretch{1}
    \caption{Watermark robustness against various video-level and model-level attacks.}
 \begin{tabular}{c|c|ccccccccccccc} 
    \toprule
     \multirow{3}{*}{\textbf{Method}} & \multirow{2}{*}{\textbf{Payload}} & & Non-  & Motion & H.264/ & Frame  & Gaussian & Gaussian & Scale & Brightness & Contrast  & Overlay & \multirow{3}{*}{\shortstack{Quanti-\\zation}} & Prun-\\
     & & \multirow{-2}{*}{\textbf{Metrics}/} & Attack & Blur & AVC  & Average  & Noise & Blur  & Adjust & Adjust & Adjust  & Emoji & & ing\\
    & (bits) & \textit{Factor} &  & std=5 & CRF=27 & $N$=4 & $\sigma$=0.02 & std=5 & $p$=0.75 & $f$=0.5 & $f$=0.5 & 25$\times$25px & & $p$=0.2\\
    \midrule
    
    \multicolumn{15}{c}{Watermarking Wan model~\cite{wan2.1}}\\
    \midrule
    
     \multirow{1}{*}{\shortstack{Stable Sig.~\cite{fernandez2023stable}}} & \multirow{1}{*}{48} & Bit Acc & 1.00 & 0.46 & 0.66 & 0.93 & 1.00 & 0.46 & 0.89 & 0.97 & 0.98 & 1.00 & 0.82 & 0.76\\
     
    \midrule
    
     \multirow{1}{*}{\shortstack{WOUAF~\cite{kim2024wouaf}}} & \multirow{1}{*}{32} &  Bit Acc & 0.72 & 0.60 & 0.53 & 0.71 & 0.62 & 0.60 & 0.62 & 0.78 & 0.74 & 0.71 & 0.62 & 0.59\\

    \midrule

     \rowcolor{gray!8} & & Bit Acc & \textbf{1.00} & \textbf{0.94} & \textbf{0.74} & \textbf{0.95} & \textbf{1.00} & \textbf{0.94}  & \textbf{1.00} & \textbf{1.00} & \textbf{0.98} & \textbf{1.00} & \textbf{0.96} & \textbf{0.94}\\
     \rowcolor{gray!8} & & TPR$_{1}$ &\textbf{1.00}&\textbf{1.00}& 0.71 &\textbf{1.00}&\textbf{1.00}&\textbf{1.00}  &\textbf{1.00}&\textbf{1.00}&\textbf{1.00} & 1.00 & \textbf{1.00} & \textbf{1.00}\\
     \rowcolor{gray!8} & & FPR$_{1}$ & 0.00 & 0.00 & 0.00 & 0.00 & 0.00 & 0.00 & 0.00 & 0.00 & 0.00 & 0.00 & 0.00 & 0.00\\
     \rowcolor{gray!8} & & TPR$_{20}$ &\textbf{1.00}&\textbf{1.00}&\textbf{1.00}&\textbf{1.00}&\textbf{1.00}&\textbf{1.00}&\textbf{1.00}&\textbf{1.00}&\textbf{1.00}& 1.00 & \textbf{1.00} & \textbf{1.00}\\
     \rowcolor{gray!8} \multirow{-5}{*}{\shortstack{\textbf{\textit{Our Scheme}}}} & \multirow{-5}{*}{\textbf{96}} & FPR$_{20}$ & 0.00 & 0.00 & 0.00 & 0.00 & 0.00 & 0.00 & 0.00 & 0.00 & 0.00 & 0.00 & 0.00 & 0.00\\

    \midrule
    \multicolumn{15}{c}{Watermarking Open-Sora model~\cite{zheng2024open}}\\
    \midrule
    
     \multirow{1}{*}{\shortstack{Stable Sig.~\cite{fernandez2023stable}}} & \multirow{1}{*}{48} & Bit Acc & 1.00 & 0.96 & 0.93 & 0.85 & 1.00 & 0.96 & 1.00 & 0.86 & 0.86 & 1.00 & 0.84 & 0.79\\
     
    \midrule
    
     \multirow{1}{*}{\shortstack{WOUAF~\cite{kim2024wouaf}}} & \multirow{1}{*}{32} &  Bit Acc & 0.75 & 0.56 & 0.66 & 0.66 & 0.64 & 0.56 & 0.63 & 0.78 & 0.71 & 0.75 & 0.64 & 0.60\\

    \midrule

     \rowcolor{gray!8} & & Bit Acc & \textbf{1.00} & \textbf{1.00} & \textbf{0.95} & \textbf{0.99} & \textbf{1.00} & \textbf{1.00} & \textbf{1.00} & \textbf{1.00} & \textbf{0.99} & \textbf{1.00} & \textbf{0.97} & \textbf{0.95}\\
     \rowcolor{gray!8} & & TPR$_{1}$ &\textbf{1.00}&\textbf{1.00}&\textbf{1.00}&\textbf{1.00}&\textbf{1.00}&\textbf{1.00}&\textbf{1.00}&\textbf{1.00}&\textbf{1.00}& 1.00 & \textbf{1.00} & \textbf{1.00}\\
     \rowcolor{gray!8} & & FPR$_{1}$ & 0.00 & 0.00 & 0.00 & 0.00 & 0.00 & 0.00 & 0.00 & 0.00 & 0.00 & 0.00 & 0.00 & 0.00\\
     \rowcolor{gray!8} & & TPR$_{20}$ &\textbf{1.00}&\textbf{1.00}&\textbf{1.00}&\textbf{1.00}&\textbf{1.00}&\textbf{1.00}&\textbf{1.00}&\textbf{1.00}&\textbf{1.00}& 1.00 & \textbf{1.00} & \textbf{1.00}\\
     \rowcolor{gray!8} \multirow{-5}{*}{\shortstack{\textbf{\textit{Our Scheme}}}} & \multirow{-5}{*}{\textbf{96}} & FPR$_{20}$ & 0.00 & 0.00 & 0.00 & 0.00 & 0.00 & 0.00 & 0.00 & 0.00 & 0.00 & 0.00 & 0.00 & 0.00\\

     \bottomrule
 \end{tabular}

\vspace{1.5mm}
\begin{minipage}{.95\linewidth}
\scriptsize
\textit{Note:} Details and parameters of the video-level attacks are provided in Table~\ref{tab:VidMark performance}. The model-level distortions applied to the watermarked VGM are defined as follows. \textbf{Quantization}: reducing the numerical precision of the model weights (e.g., to 8-bit); \textbf{Pruning}: sparsifying the network by setting $p$ weights to zero.
Additionally, $\text{TPR}_1$ and $\text{FPR}_1$ represent the results of true positive rate and false positive rate on the single-video scenario, while $\text{TPR}_{20}$ and $\text{FPR}_{20}$ denote the results in the multiple-video scenario with $n=20$. The corresponding thresholds applied here are detailed in Table~\ref{tab:theoretical_bounds}.
\end{minipage}

\label{tab_robustness}
\end{table*}

\begin{figure*}[t]
\centering
\includegraphics[width=.95\linewidth]{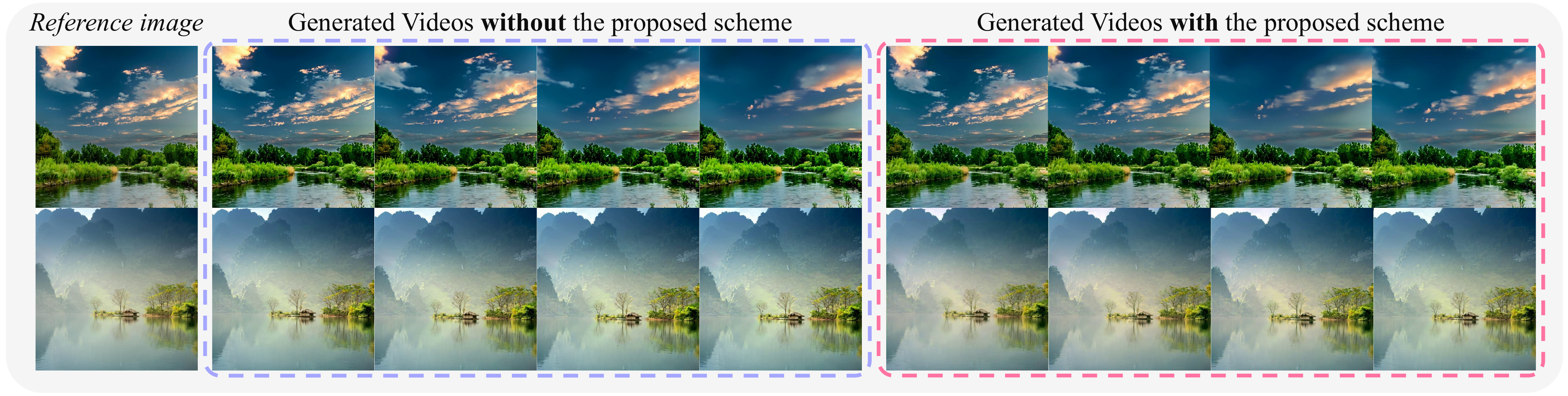}
\caption{
    I2V generation results produced by the original SVD model~\cite{blattmann2023stable} (\textbf{without} watermark) and the watermarked SVD model (\textbf{with} our proposed scheme). 
}
\label{Fig_visual_comparison2}
\end{figure*}

\subsubsection{Impact on Video Generation}
Table~\ref{tab:main table} summarizes the quantitative results of visual quality and watermark extraction accuracy in the non-attack scenario. As shown, Stable Sig.~\cite{fernandez2023stable} achieves perfect extraction accuracy (i.e., Bit Acc = 1.0) but at the cost of severe video quality degradation, with the FVD increasing to 191.07 and 751.50 on Wan and Open-Sora, respectively. Although WOUAF~\cite{kim2024wouaf} preserves good visual quality, it cannot guarantee reliable watermark extraction, with a Bit Acc of only 0.72 to 0.76. In comparison, our scheme achieves a Bit Acc of 1.00 with the largest payloads (96 bits for Wan and Open-Sora, 64 bits for SVD), while maintaining video quality comparable to the non-watermarked baselines.

To further evaluate the actual video generation quality, Fig.~\ref{Fig_visual_comparison} provides a qualitative comparison of the generated videos. 
As shown in Fig.~\ref{Fig_visual_comparison}, WOUAF exhibits severe color shifts in the generated videos, despite achieving relatively favorable scores on several quantitative metrics in Table~\ref{tab:main table}.
We speculate that this difference is due to overfitting of the VAE encoding-decoding process by the embedding mechanism of WOUAF~\cite{kim2024wouaf}. Specifically, it is highly sensitive to the inherent distributional bias between the latents directly encoded by the VAE encoder and the latents synthesized in the VGM generation backbone. Therefore, during actual generation, this bias is amplified, ultimately leading to significant color degradation. Conversely, our scheme effectively avoids such issue, generating videos that are visually identical to the original Wan model. Furthermore, the corresponding difference frames in Fig.~\ref{Fig_visual_comparison} confirm that our scheme achieves near-zero visual degradation compared to the baseline Wan model. Similarly, for the image-to-video (I2V) synthesis, Fig.~\ref{Fig_visual_comparison2} illustrates that the watermarked SVD model maintains the original generation quality and produces visually identical videos to the original SVD model. Overall, these results demonstrate that our proposed scheme introduces negligible impact on the original video generation quality of the VGMs.

\subsubsection{Watermark Robustness}
Table~\ref{tab_robustness} details the robustness of the embedded watermarks against various video-level and model-level attacks, evaluated on the two recent frameworks (i.e., Open-Sora~\cite{zheng2024open} and Wan~\cite{wan2.1}).
Under these attacks, the baseline methods fail to maintain reliable extraction accuracy. For example, on the Wan model~\cite{wan2.1}, the Bit Acc of Stable Sig.~\cite{fernandez2023stable} drops to 0.46 under motion blur and 0.66 under H.264 compression, while WOUAF~\cite{kim2024wouaf} generally fails across all attacks, with its Bit Acc dropping to around 0.60. In contrast, our scheme stably maintains a high Bit Acc in most attack scenarios. Notably, our scheme also demonstrates strong robustness against model-level attacks, maintaining high Bit Acc (above 0.94) even after model quantization and pruning.

Furthermore, the results in Table~\ref{tab_robustness} validate the effectiveness of the proposed verification framework. In the most severe distortion scenario, heavy video degradation inevitably affects the individual watermark extraction, causing $\text{TPR}_1$ to drop to 0.71. However, by querying $n=20$ videos in the multiple-video scenario, {$\text{TPR}_{20}$ remains stable at 1.00. Moreover, across all attacks, the false positive rates ($\text{FPR}_1$ and $\text{FPR}_{20}$) remain strictly at 0.00. This demonstrates that the aggregated verification strategy effectively overcomes the inherent uncertainties of single-video extraction, ensuring reliable model ownership verification even under extreme attacks.

\begin{table}[t]
    \centering
    \setlength{\tabcolsep}{1.5mm}{}
    \renewcommand\arraystretch{1}
    \caption{
        Comparison of different methods regarding complexity. The VGM selected here is SVD model~\cite{blattmann2023stable}. 
    }
    \begin{tabular}{l|cc} 
    \toprule
        \multirow{1}{*}{\textbf{Method}}  & \multirow{1}{*}{{\shortstack{Memory cost (MiB)}}} & \multirow{1}{*}{{\shortstack{Inference time (s)}}} \\
    \midrule
        \textit{Non-watermarked} & \textit{68752} & 97.50\\
        Stable Sig. \cite{fernandez2023stable}  & 77476  & 97.51\\
        WOUAF~\cite{kim2024wouaf} & 79912 & 98.58\\
    \rowcolor{gray!8} 
        \textbf{\textit{Our Scheme}}& \textbf{70218}  &  97.50\\
    \bottomrule
\end{tabular}
\label{tab:Complexity}
\end{table}

\subsubsection{Complexity}
To evaluate computational efficiency, Table~\ref{tab:Complexity} compares the memory costs of different schemes during SVD model~\cite{blattmann2023stable} fine-tuning and the inference time of the watermarked model and the baseline SVD model during I2V generation. As seen, in the fine-tuning stage, the memory overhead introduced by our scheme can be ignored, only slightly increasing the memory usage of the non-watermarked baseline model from 68,752 MiB to 70,218 MiB, whereas Stable Sig.~\cite{fernandez2023stable} and WOUAF~\cite{kim2024wouaf} require higher resources (77,476 MiB and 79,912 MiB, respectively). Furthermore, during I2V generation, our watermarked model requires 97.51 seconds for inference, which aligns with the original model (97.50 seconds) and avoids the slight computational delay introduced by WOUAF (98.58 seconds). Overall, these results confirm that the proposed scheme is lightweight and ensures that no additional computational overhead is incurred during the actual video generation process.

\begin{table}[t]
    \centering
    \setlength{\tabcolsep}{2.2mm}{}
    \renewcommand\arraystretch{1.2}
    \caption{
        Performance of watermarking Wan~\cite{wan2.1} when using different watermark decoders.
    }
    \begin{tabular}{lccc} 
    \toprule
        Used Watermark Decoder & PSNR $\uparrow$ & LPIPS $\downarrow$ & Bit Acc $\uparrow$ \\
    \midrule
        REVMark~\cite{zhang2023novel} & 30.66  &  0.092 & 0.998   \\
        \rowcolor{gray!8} 
        VidMark  & \textbf{34.61}  & \textbf{0.059} & \textbf{0.999} \\
    \bottomrule
    \end{tabular}
    \label{tab:ablation study on watermark decoder}
\end{table}

\subsubsection{Impact of the VidMark Decoder}
\label{sec5_3_ablation_studies_of_proposed_watermarking_scheme}
To verify the necessity of the proposed VidMark in VGM watermarking, we replace the VidMark decoder with the decoder of REVMark~\cite{zhang2023novel} during the fine-tuning process. Table~\ref{tab:ablation study on watermark decoder} shows the video quality and watermark extraction results on the Wan model~\cite{wan2.1}. Both decoders achieve high watermark extraction accuracy with a Bit Acc near 1.00. However, using the REVMark decoder causes a noticeable drop in visual quality, resulting in a PSNR of 30.66 dB and an LPIPS of 0.092. In contrast, our VidMark decoder maintains high visual fidelity, achieving a PSNR of 34.61 dB and an LPIPS of 0.059. These results prove that the specific design of VidMark, including the two-scale DWT and GTAB, helps the VGM learn better watermark embedding behaviors, thereby balancing the trade-off between robust watermark extraction and visual quality preservation.

\section{Conclusion}
\label{sec6_conclusion}

In this paper, we propose an efficient in-generation watermarking scheme to safeguard the IPR of VGMs. First, we design VidMark, a lightweight video watermarking network that integrates two-scale DWT and GTAB to achieve robust and imperceptible watermark embedding in high-resolution videos. Second, we present a decoder-guided fine-tuning framework that selectively optimizes the output-proximate layers of the VGM's VAE decoder, enabling the model to inherently synthesize watermarked outputs without compromising the original generation quality. Third, we introduce a rigorous verification framework based on statistical hypothesis testing, which enables reliable model ownership verification when multiple generated videos are available, even under severe distortions. Extensive experiments on state-of-the-art VGMs, including SVD~\cite{blattmann2023stable}, Open-Sora~\cite{zheng2024open}, and Wan~\cite{wan2.1}, demonstrate that our scheme consistently achieves 100\% verification accuracy, preserves visual fidelity, and exhibits excellent robustness against various video-level and model-level attacks. 

\bibliographystyle{IEEEtran}
\bibliography{ref}

\end{document}